%% file: root.tex
\documentclass[a4paper]{svproc}
\pdfoutput=1
\usepackage{graphicx, wrapfig}
\BeforeBeginEnvironment{wrapfigure}{\setlength{\intextsep}{0pt}}
\usepackage{amsmath, amssymb, amsfonts, siunitx}
\usepackage{mathtools}
\DeclarePairedDelimiterX{\norm}[1]{\lVert}{\rVert}{#1}
\usepackage[dvipsnames]{xcolor}
\usepackage{tikz,amsmath, amssymb,bm,color}
\usepackage{graphicx}
\usepackage{pgfplots}
\usepackage{pgfgantt}
\usepackage{pdfpages}
\usepackage{caption}
\usepackage{subcaption}
\usepackage{svg}
\usepackage{float}
\usepackage{enumitem}
\usepackage{setspace}
\usepackage{array}
\usetikzlibrary{shapes,arrows,patterns,3d}
\usetikzlibrary{pgfplots.units}
\usetikzlibrary{pgfplots.statistics} 
\usetikzlibrary{chains}

\usepackage{url}

\newcommand{\vectorvar}[1]{\Vec{#1}}

\newcommand{\rulesep}{\unskip\ \vrule\ }

\begin{document}
\mainmatter              % start of a contribution
\title{Modeling Aggregate Downwash Forces for Dense Multirotor Flight}
% "Aggregate Downwash Forces.."
% "
%
\titlerunning{Modeling Aggregate Downwash Forces}  % abbreviated title (for running head)
\author{Jennifer Gielis
\and Ajay Shankar
\and Ryan Kortvelesy
\and Amanda Prorok
}
\authorrunning{Jennifer Gielis et al.} % abbreviated author list (for running head)
\tocauthor{Jennifer Gielis, Ajay Shankar, Ryan Kortvelesy, Amanda Prorok}
\institute{Department of Computer Science \& Technology, University of Cambridge, UK\\
\email{\{jag233, as3233, rk627, asp45\}@cl.cam.ac.uk},
}

\maketitle              % typeset the title of the contribution

\begin{abstract}
%Dense formation flight with multiple multirotor aerial vehicles is a nature-inspired flight regime that employs various capabilities in collective intelligence, planning, and art. \amanda{dont understand the preceding sentence.}
Dense formation flight with multirotor swarms is a powerful, nature-inspired flight regime with numerous applications in the real-world.
However, when multirotors fly in close vertical proximity to each other, the propeller downwash from the vehicles can have a destabilising effect on each other.
Unfortunately, even in a homogeneous team, an accurate model of downwash forces from one vehicle is unlikely to be sufficient for predicting aggregate forces from multiple vehicles in formation.
In this work, we model the interaction patterns produced by one or more vehicles flying in close proximity to an ego-vehicle.
We first present an experimental test rig designed to capture 6-DOF exogenic forces acting on a multirotor frame.
We then study and characterize these measured forces as a function of the relative states of two multirotors flying various patterns in its vicinity.
Our analysis captures strong non-linearities present in the aggregation of these interactions.
Then, by modeling the formation as a graph, we present a novel approach for \textit{learning} the force aggregation function, and contrast it against simpler linear models.
Finally, we explore how our proposed models generalize when a fourth vehicle is added to the formation.
\end{abstract}

\input{01-motivation.tex}
\input{02-technical.tex}
\input{03-results.tex}
\input{04-insights.tex}
\bibliographystyle{spmpsci_unsrt}
\vspace{-5ex}
\begin{singlespace}
{\tiny
\bibliography{ref,aj_refs}
}
\end{singlespace}
\end{document}

%% file: 01-motivation.tex
\section{Motivation}
Formation flights with aerial swarms push the limits of sensing, prediction, control and coordination.
They also present a unique, highly dynamic setting for
art~\cite{du2019fast},
collective intelligence, and
force multiplication, where the team can collectively span a very large volume and track multiple moving targets with greater degree of redundancy~\cite{chung2018survey}.
The next frontier we are exploring in this regime is \textit{dense} flight, where multirotors frequently fly within 1-3 body-lengths of each other.
This might often be a mission requirement (e.g., fly through a passage).

A key physical limitation in realising this arises due to the turbulent aerodynamic interactions between two multirotors flying in close proximity to each other, especially when aligned directly beneath one another.
It is crucial to have some form of a model for these `exogenic' forces, since reactive controllers are not able to reject them without first observing the disturbance.
These interactions are typically hard to model even when there is only one neighbour in the vicinity~\cite{smith2023so}.
Nevertheless prior work has shown success in constructing a single-neighbour model either analytically~\cite{jain2019modeling} or in a data-driven fashion~\cite{smith2023so}.

\begin{figure}[t]
    \centering
    \includegraphics[width=0.42\textwidth]{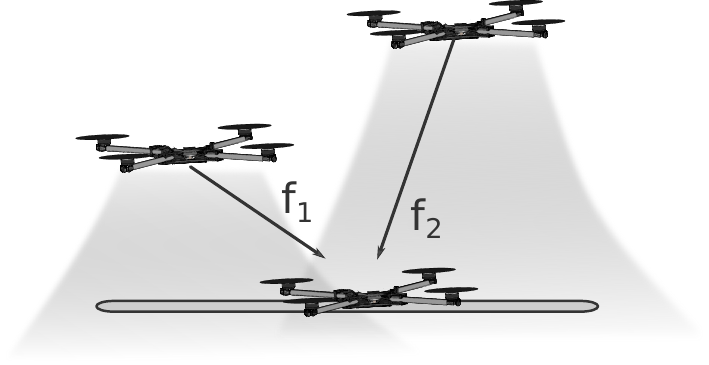}
    \rulesep
    \includegraphics[width=0.42\textwidth]{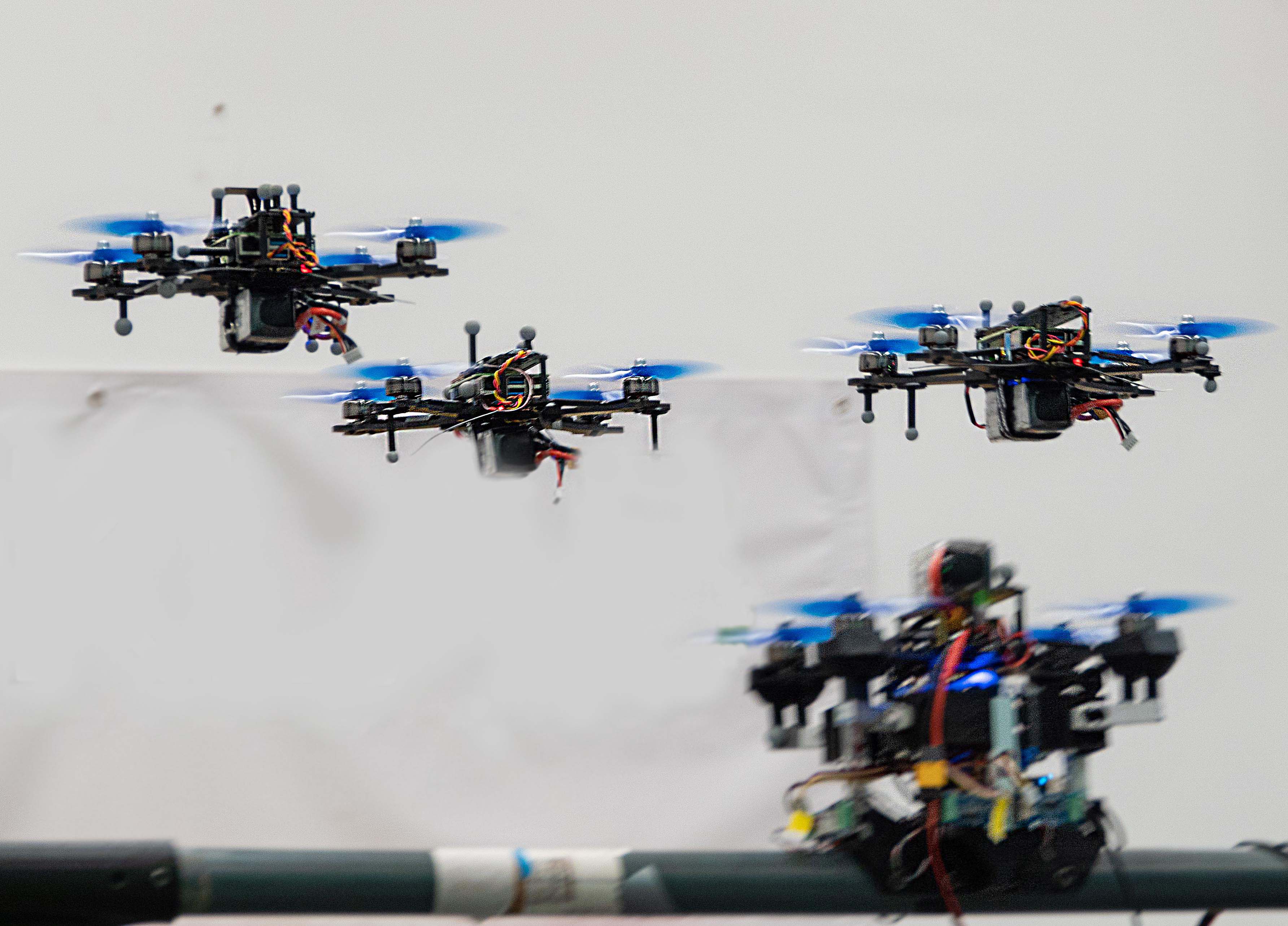}
    \caption{An illustration of the problem we are studying (left), the load-stand design (center), and a still from flight tests with multiple vehicles in formation over a sufferer fixed to the load-stand (right).}
    \vspace{-4ex}
    \label{fig:LoadStandSetup}
\end{figure}

A na\"ive approach would then query this single-neighbour downwash-force model, $\vectorvar{f}(\boldsymbol{x}_j)$, with $\boldsymbol{x}_j$ denoting the state relative to neighbour $j$, in a pairwise manner for all neighbouring multirotors $j = \{0 \ldots K\}$ to obtain a simple downwash predictive model of the total force experienced by the ego-vehicle (referred to as the ``sufferer" in this work), 
$f^i = \Sigma_{j=0}^K \vectorvar{f}(\boldsymbol{x}_j)$.
%, where $f^i$ is the force vector experienced by the ego multirotor, which we refer to as the \textit{sufferer}.
This makes the assumption that downwash forces from the $K$ neighbours are additive, or at least, linearly summable.
While the assumption \textit{might} hold for a variety of flight regimes/formations, there is a dearth of empirical studies that explore this further.
In particular, we lack a systematic means for identifying formations that \textit{necessitate} more complex models.

\textbf{Problem Statement.}
The objective in this work is to conduct an investigation into downwash forces in various multi-vehicle flight regimes, and then empirically model the force aggregation for $K$ multirotors around a sufferer.

Towards this end, we address two key questions:
\begin{description}[leftmargin=0pt,topsep=0pt]
\item \textbf{Q1:} Given a single-neighbour downwash model, $\vectorvar{f}$, can we perform a \textit{pairwise} linear combination of functions $\vectorvar{f}$ for each neighbour to obtain an accurate estimate of collective force from $K$ neighbours as experienced by a sufferer?
\item \textbf{Q2:} In cases where this linear combination may be insufficient, does a learnt nonlinear model architecture provide sufficient expressivity in order to capture complex aerodynamic interactions at scale?
\end{description}

\subsubsection*{Related Work}
Previous work has addressed the question of multirotor aerodynamic interactions \cite{yeo2015empirical}, which often includes complex methods and equipment \cite{carter2021influence}. Zhang et al. \cite{zhang2018} perform a numerical analysis of the downwash a single vehicle generates and show the general shape of airflow.
Shi et al. \cite{shi2020neural} implicitly learn the impact of neighbour's presence with small quadrotors, showing some improvement in trajectory tracking through a learning-based downwash prediction.

Computational Fluid Dynamics (CFD) has also commonly been used to model the interactions of single vehicles, and are often verified on test vehicles \cite{Diaz2018} \cite{GUO2020105343}. The computational requirements of this approach are impractical to scale up resulting in a dearth of studies that cover multi-quadrotor swarms.

Li et al. \cite{li2023nonlinear} is a recent approach similar to ours that corrects for single quadrotor downwash using a learning-based method with load cell collected ground truth data. Our approach furthers this by accounting for up to 3 neighbour multirotors, a 6-DOF Load Stand, and several modelling approaches. Together these features allow us to perform an in-depth analysis of the airflow around formations of multirotors in flight, as well as produce deploy-able predictive models that can be used to improve positional accuracy in dense swarms.

To our knowledge, no previous work has empirically quantified the characteristics of downwash for a range of specific dense swarm formations to this depth.

%% file: 02-technical.tex
\section{Technical Approach}

In Section~\ref{sec:experimental}, we describe our experimental setup for accurately measuring ground-truth exogenic forces.
Here, we present three modelling approaches to \textit{predict} the forces due to multiple vehicles flying in formation:

\subsubsection{Linear Model} This class assumes that the net force is equal to the sum of forces due to individual vehicles. The force prediction is computed as a sum of the single-multirotor models $\psi: \mathbb{R}^{d_\mathrm{in}} \to \mathbb{R}^{d_\mathrm{out}}$, which operates over the relative state $\boldsymbol{x}_j - \boldsymbol{x}_i$. The model is given by:
{\small
\begin{equation}
    \hat{f}_i = \sum_{j=1}^K \psi(\boldsymbol{x}_j - \boldsymbol{x}_i).
\end{equation}
}
\vspace{-3ex}

\subsubsection{Nonlinear Model} Based on Deep Sets \cite{deepsets}, this class enables universal set function approximation. Through a composition of functions $\phi: \mathbb{R}^{d_\mathrm{in}} \to \mathbb{R}^h$ and $\Phi: \mathbb{R}^h \to \mathbb{R}^{d_\mathrm{out}}$, it can consider not only pairwise relationships, but $K$-wise relationships. Our model takes the form:
\begin{equation}
    \hat{f}_i = \Phi \left( \sum_{j=1}^K \phi(\boldsymbol{x}_j - \boldsymbol{x}_i) \right).
\end{equation}
\vspace{-1ex}

Each model has its own advantages and disadvantages. The linear model can be trained on single-multirotor data, and it provides the capacity to generalise to a larger $K$ than what was experienced in the training data.
%On the other hand, the nonlinear model experiences an input with a different number of vehicles $K$ as out-of-distribution.
To a (learnt) non-linear model, on the other hand, an input with a different $K$ is completely out of distribution.
However, due to the chaotic nature of fluid mechanics, the linear extrapolation used by the linear model cannot necessarily be assumed. It is possible that a set of $K$ vehicles exhibits significantly different dynamics than a set of $K+1$ vehicles, thus necessitating a nonlinear model. As a universal set function approximator, our nonlinear model can model behaviour between cliques of multiple multirotors, and can learn distinct behaviour for different numbers of vehicles $K$. This additional representational capacity enables the nonlinear model to make more accurate predictions.

%% file: 03-results.tex
\section{Experimental Setup}
\label{sec:experimental}
Our experimental process has two key elements: first, a bespoke load-stand setup that allows us to accurately measure 6 degree-of-freedom (6-DOF) forces acting on a multirotor, and second, exploration strategies that capture these forces over various trajectories flown by various formations.

\begin{figure}[t]
    \centering
    \resizebox{1.0\columnwidth}{!}{%
    \includegraphics[width=0.48\textwidth]{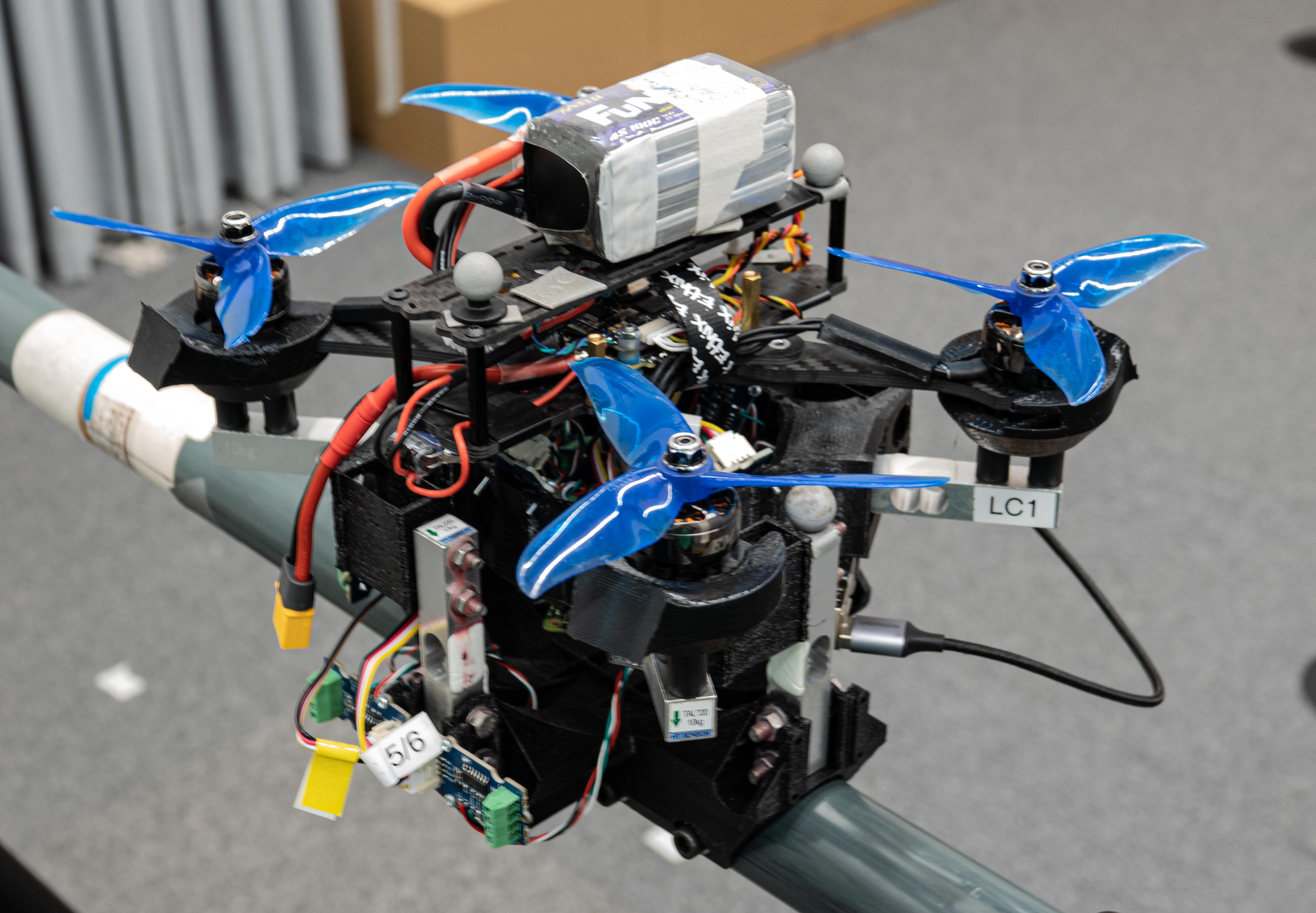}
    \rulesep
    \includegraphics[width=0.48\textwidth]{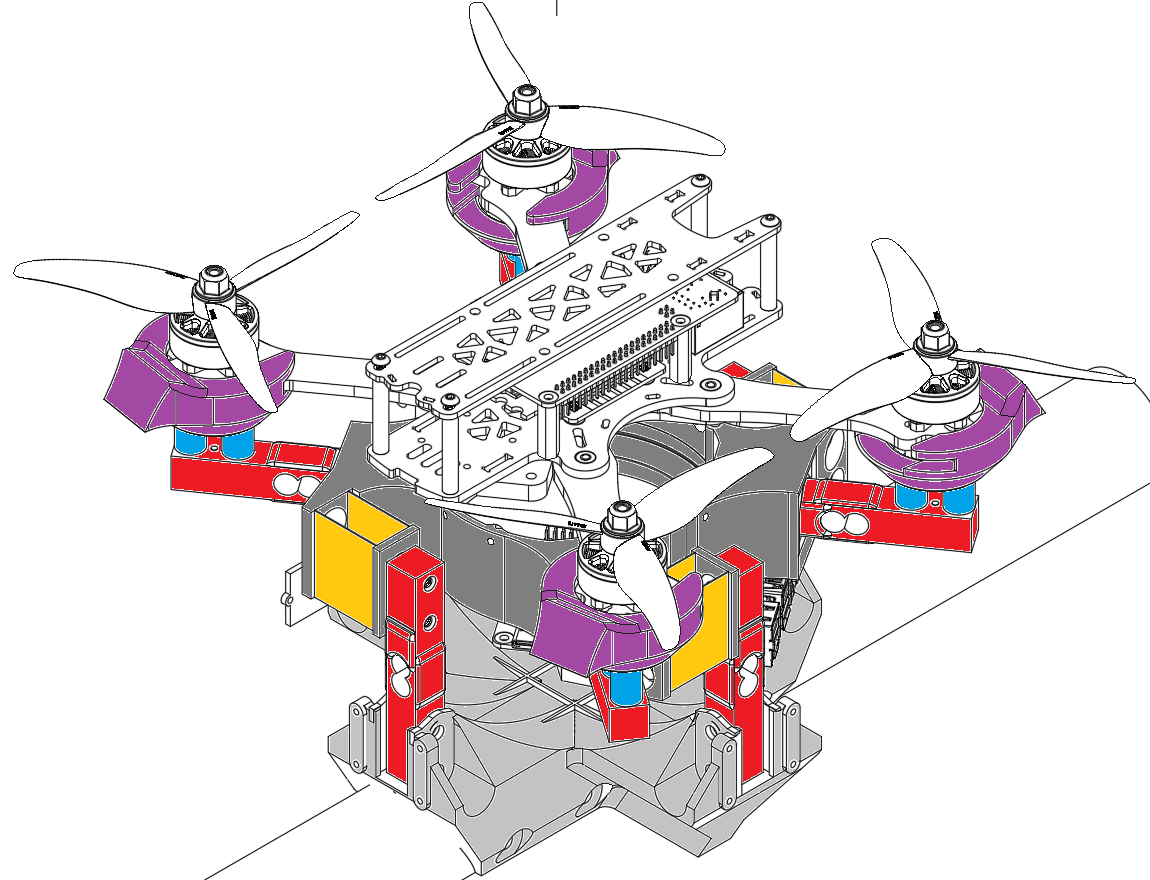}
    }
    \caption{A diagram representing our 6-DOF load-stand setup. Coloured components can be identified as follows. White: Multirotor, Purple: Couplers, Blue: Rubber Dampers, Red: Load Cells, Yellow: Selectively compliant transfer elements, Dark grey: Lateral transfer element, Light grey: Pipe anchor}
    \label{fig:LoadStandDesign}
\end{figure}

\subsubsection{Load-Stand for Ground Truth Data.}
In order to evaluate any of the models described above, we first require a means to capture the forces acting on a multirotor frame.
This measurement process is inherently non-trivial, and requires some essential considerations to be made.
First, it must be fast and decoupled in all axes to be able to capture transient effects as the neighbours in a formation fly by.
Second, it must be free of the biases from the employed flight controller and its response characteristics.
Finally, for the measurements to be meaningful, the frame of the sufferer must be in a ``flying-like" state, i.e., its propellers must be spinning and generating thrust so that the airflows mix as they would with an unconstrained hovering multirotor.

To satisfy these objectives, we design and build a novel load-stand setup, shown in Fig. \ref{fig:LoadStandDesign}, comprised of eight load-cells that register shearing forces, and mechanically constrain the sufferer to it using couplers at each motor joint.

Four of these cells, connected via a pair of vibration-damping rubbers to the couplers, measure shear forces in the vertical axis. Our experimental results are reported in the standard aerospace North-East-Down (NED) frame convention, where the vertical is the D-axis. Readings from these cells are combined in order to compute the overall D-axis force, as well as the pitch and roll torques exerted upon the platform.

This set of motor boom load-cells are connected at their base to a lateral transfer element which in turn connects to another set of four load cells that measure lateral forces, with two measuring in each of the N and E-axes.
The lateral cells are connected via flexible elements that are mechanically compliant in the lateral axis perpendicular to
%the attached load
a
cell's measurement axis, ensuring forces are recorded accurately from the complimentary cells reading in that axis.
These 4 lateral cells are read together to produce N and E-axis forces as well as yaw torques.
The overall arrangement of 8 load cells allows 6 DOF measurements of forces acting upon the platform under measurement.

Finally, the load-stand as a whole is mounted on a horizontal pole and held \SI{1.4}{m} above the ground, a height chosen to avoid ground interaction aerodynamic effects, with the overall lateral cross section of the stand designed to avoid components intruding into each propeller's downwash field.

The load cells are shear-beam type with a \SI{10}{kg} capacity. We read them through a set of HX711 differential amplifier and ADC ASIC packages. These devices are configured to run at an \SI{80}{Hz} rate which we read back through GPIO pins on a Raspberry Pi 4B that is integrated into the load-stand below the lateral transfer plate. This configuration limits interference from external wiring.

This system performs a self-calibration routine at the beginning of each experiment; first the constrained sufferer sets its motor thrust to hover thrust (thus zeroing the D-axis forces on the stand), and then the load-stand software reads the forces on all cells and uses those readings as a new zero point. The computation of this offset is trivial due to the highly linear nature of our load cells. This process ensures that forces presented are similar to those acting upon an unconstrained vehicle in hover, and also ensures power cable connections into the load-stand do not produce varying biases between experiments. Additionally, factors that vary between experiments such as ambient air pressure and temperature are effectively corrected for.

\subsubsection{Exploration Strategy.}
\begin{figure}[tb]
    \centering
    \resizebox{0.98\columnwidth}{!}{%
\begin{tikzpicture}
	\tikzstyle{building} = [fill=black!20];
	
	\definecolor{refcolor}{HTML}{bdbdbd}
	\definecolor{boxcolor}{HTML}{f0f0f0}
	\tikzstyle{droneprop} = [fill=refcolor!50, fill opacity=100];
	\tikzstyle{velcolour} = [color=cyan]
	\tikzstyle{vel} = [-latex, dashed, ultra thick, velcolour];
	
	\tikzset{
		drone/.pic = {
			\draw[thick] (-0.2,0.1) node (v4) {} -- (0.4,-0.5) node (v2) {};
			\draw[thick] (-0.2,-0.5) node (v1) {} -- (0.4,0.1) node (v3) {};
			\draw[fill]  (0,0) rectangle (0.2,-0.4);
			\draw[droneprop]  (v1) ellipse (0.2 and 0.2);
			\draw[droneprop]  (v2) node (v5) {} ellipse (0.2 and 0.2);
			\draw[droneprop]  (v3) ellipse (0.2 and 0.2);
			\draw[droneprop]  (v4) ellipse (0.2 and 0.2);
			\draw[fill] (v1) ellipse (0.05 and 0.05);
			\draw[fill] (v2) ellipse (0.05 and 0.05);
			\draw[fill] (v3) ellipse (0.05 and 0.05);
			\draw[fill] (v4) ellipse (0.05 and 0.05);
		}
	}
	
	\tikzset{
		droneWvel/.pic = {
			\filldraw[opacity=0,fill=red!50, fill opacity=0.5] (0.1, -0.2) circle (10mm);
			\path (0,0) pic{drone};
			\draw [vel] (0.1,0) -- (0.1,1);
		}
	}
	
	\tikzstyle{wifibar} = [line width=4pt, line cap=round];
	\tikzset{
		wifibars/.pic = {
			\draw[wifibar] (0,0) arc (0:80:0.4);
			\draw[wifibar] (0.2,0) arc (0:80:0.6);
			\draw[wifibar] (0.4,0) arc (0:80:0.8);
		}
	}
	
	\path [rotate=90,scale=0.9,transform shape,color=teal!90] (2.2222,2.4) pic{drone};
	\path [rotate=90,transform shape] (1.2,1.2) pic{droneWvel};
	\path [rotate=90,transform shape] (2.8,1.2) pic{droneWvel};

	\tikzstyle{commsarrow} = [-latex, thick, dashed];
	\tikzstyle{commsarrowinter} = [thick, dashed];
	\tikzstyle{commsarrowfaded} = [-latex,thick, dashed, color=black!10];

	\node [fill=white, draw=black] at (-1.3,3.8) {\large\emph{Side by Side}};

	\path [rotate=90,scale=0.9,transform shape,color=teal!90] (2.1778,-2.4) pic{drone};
	\path [rotate=90,transform shape] (2,-3) pic{droneWvel};
	\path [rotate=90,transform shape] (2,-1.2) pic{droneWvel};

	\node [fill=white, draw=black] at (2.2,3.8) {\large\emph{Leader Follower}};
	
	\draw (0.1,3.6) -- (0.1,0.4);
	
	\path [rotate=90,scale=0.9,transform shape,color=teal!90] (2.3499,-5.1668) pic{drone};
	\path [rotate=90,transform shape] (1.6,-5.9) pic{droneWvel};
	\path [rotate=90,scale=1.1,transform shape] (2.3182,-5.2728) pic{droneWvel};
	
	\node [fill=white, draw=black] at (5.9,3.8) {\large\emph{Stack}};

	\draw (4.3,3.6) -- (4.3,0.4);
	
	\path [rotate=90,scale=0.9,transform shape,color=teal!90] (2.1,-10.2) pic{drone};
	\path [rotate=90,transform shape] (1.1,-9.8) pic{droneWvel};
	\path [rotate=90,transform shape] (1.9,-8.3) pic{droneWvel};
	\path [rotate=90,transform shape] (2.7,-9.8) pic{droneWvel};
	\node [fill=white, draw=black] at (9.1,3.8) {\large\emph{3 Hybrid}};
	
	\draw (7.2,3.6) -- (7.2,0.4);
\end{tikzpicture}
    }
    \caption{An illustration of the three main flight patterns we consider, plus a hybrid case for 3 vehicles. The green vehicle is the load-stand mounted (fixed) sufferer, and arrows indicate vehicle direction of travel.}
    \vspace{-2ex}
    \label{fig:formation-patterns}
\end{figure}
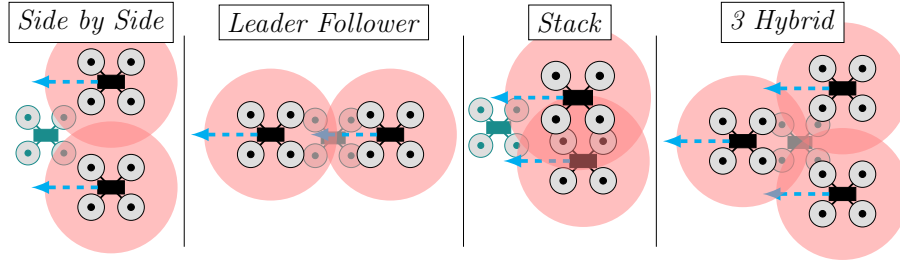

In order to practically explore the possible set of state configurations that are most likely to generate complex downwash interactions with two mobile vehicles, we define a set of three formation patterns as exemplars of extreme close flight, and then investigate interactions with the load-stand multirotor over the state-space, defined below. We show these visually in Fig.~\ref{fig:formation-patterns} and describe them as follows:

\begin{itemize}[itemsep=0pt,labelindent=*]
    \item \textit{Side by Side}: Multirotors fly a formation perpendicular to the direction of travel on the lateral plane. Downwashes reach the sufferer simultaneously.
    \item \textit{Leader Follower}: Multirotors fly a formation parallel to the direction of travel on the horizontal plane. Downwashes interacts sequentially with the sufferer.
    \item \textit{Stack:} Multirotors fly in distinct horizontal planes, with lateral separations such that each multirotor is interacting with the downwash of those higher.
\end{itemize}

As a result of, and as an extension to, the 2-multirotor formations, we investigated two 3-multirotor formations. Firstly, an extension of the \textit{Leader Follower} case with a third vehicle, and secondly a combination of the \textit{Side by Side} and \textit{Leader Follower} cases where there are three vehicles in a equilateral triangle on a single lateral plane.

Our dataset is comprised of a dense collection of samples obtained by flying these formations in a grid-like manner over the state-space of interest.
%We collect the our dataset
This is done
over a \SI{5.6}{m^3} volume
(within a larger space of \SI{96}{m^3}), %4m N, 6m E, 4m D?
exploring a square area \SI{2}{m} on a side laterally from the load-stand, and 1.4m vertically above it. 
All trajectories are flown at \SI{0.5}{m/s}, roughly aligned with the E-axis, with 30-40 legs being flown for each experiment.
The individual trajectory legs are laid in straight line segments passing through the state-space to minimise the confounding forces that other variable-thrust trajectories might produce.
The arena is fitted with motion-capture cameras that provide positioning data to all entities at \SI{180}{Hz}. Experimental control and logging is performed by the Raspberry Pi on the load-stand.

\section{Results}
\begin{figure}[tb]
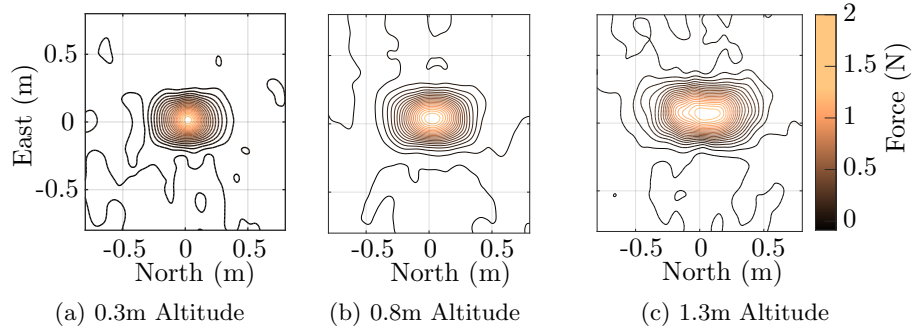

    \centering
    \begin{subfigure}[b]{0.294\textwidth}
        \centering
        \includesvg[width=\textwidth]{images/SingleLow.svg}
        \caption{0.3m Altitude}
        \label{fig:SingleContourLow}
    \end{subfigure}
    \begin{subfigure}[b]{0.28\textwidth}
        \centering
        \includesvg[width=\textwidth]{images/SingleMid.svg}
        \caption{0.8m Altitude}
        \label{fig:SingleContourMid}
    \end{subfigure}
    \begin{subfigure}[b]{0.37\textwidth}
        \centering
        \includesvg[width=\textwidth]{images/SingleHigh.svg}
        \caption{1.3m Altitude}
        \label{fig:SingleContourHigh}
    \end{subfigure}
    \caption{ D-axis forces from a single multirotor interacting with the load-stand, at varying altitudes. It can be seen that the downwash forces generally distribute more evenly as relative altitude increase, but that the width of the downwash column does not significantly expand laterally. }
    \label{fig:SingleContour}
\end{figure}

In addressing \textbf{Q1}, we evaluate whether a numeric model generated from a dataset of single neighbour flights can be na\"ively summed for neighbour counts of $K > 1$, and contrast with the predictions of the learnt models. Ground truth is obtained from load-stand readings during actual multi-vehicle formation flights.

Depending on the formation and its location in the state-space, our results of aerodynamic interactions with the load-stand
%and the various vehicle formations
generally fall into one of three categories:
%varying over the state space and formation:
\begin{enumerate}
    \item Chaotic Interactions: those that are dominated by noise;
    \item Linear Interactions: those that are well estimated by linear summation; and,
    \item Non-linear Interactions: those that can only be estimated with non-linear models.
\end{enumerate}

\subsection{Chaotic Interactions}
In general, we find that even though there are identifiable trends in data for the N and E-axis forces, as well as yaw torques, these are generally dominated by noise of various types. Noise in this context takes on several meanings.

Firstly, measurement noise; this type of noise originates within the load-stand system itself.
%Primarily this originates from
Common causes include
ADC read inaccuracy and mechanical vibrations from the constrained platform's motors, including excitation of resonances in the mount. These sources are significant, but readily quantifiable at +/- 0.05N. 

A second source of noise comes from the environmental configuration, such as objects near the load-stand. Regardless of our precautions, the presence of the load-stand setup will invariably cause some modification and redirection of the airflows\footnote{We liken this to the ``observer effect".}.
While these are held constant through all tests, we note that the combined effect of these experiment items (such as mounts, platforms, power supplies etc) do create significant effects on all vehicles.

\subsection{Linear Interactions}
\begin{wrapfigure}{r}{0.45\textwidth}
    \centering

    \begin{subfigure}[b]{0.40\textwidth}
        \centering
        \includesvg[width=\textwidth]{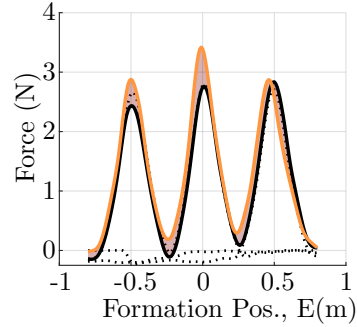}
        \vspace{-2ex}
        \caption{0.3m Relative Altitude}
        \label{fig:SxSSliceLow}
    \end{subfigure}
    \begin{subfigure}[b]{0.40\textwidth}
        \centering
        \includesvg[width=\textwidth]{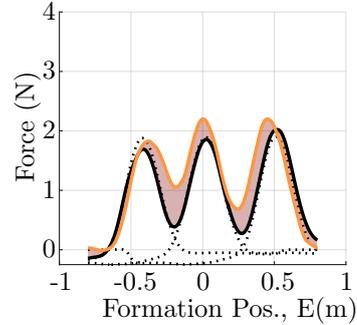}
        \vspace{-2ex}
        \caption{0.8m Relative Altitude}
        \label{fig:SxSSliceMid}
    \end{subfigure}
    \vspace{-4ex}
    \begin{subfigure}[b]{0.40\textwidth}
        \centering
        \includesvg[width=\textwidth]{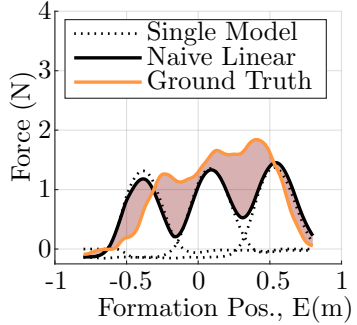}
        \vspace{-2ex}
        \caption{1.3m Relative Altitude}
        \label{fig:SxSSliceHigh}
    \end{subfigure}
    \vspace{2ex}
    \caption{Forces in the D-axis for the 3-multirotor \textit{Leader Follower} case, at varying altitudes.}
    \label{fig:SxSSlice}
    \vspace{-1ex}
\end{wrapfigure}

Linear interactions tend to occur in cases when one neighbour is significantly closer to the load-stand than others in the formation.
This is most noticeable when the formation is operating at extremely low altitude offsets (to the sufferer), such as in Fig. \ref{fig:SxSSliceLow}.
This is somewhat intuitive since, first, at low altitudes, only one neighbour can physically be in such extreme proximity at a time.
Secondly, there is no space for complex interactions in airflow patterns to occur
%between formation members
before reaching the load-stand.

%The reader may be tempted
It may be tempting
to conclude that non-linear models would be unnecessary because of this
%property of extreme close passes being largely linear,
effect,
combined with the intuition that forces from distant neighbours are likely to diminish rapidly.
Our results show that this is not the case, as can be seen in Fig. \ref{fig:SingleContour}, which shows the D-axis forces from a single neighbour. % and critically that 

We observe that the downwash field generated by the multirotor  expands very little in the directions orthogonal to the airflow direction, and that high force levels are measurable even at separations of several body-lengths (even beyond our setup's capacity to measure due to height limits). This indicates that, we could expect airflow from a distant neighbour to non-linearly interact with a proximal one with significant effect. In the following subsection, we observe that this is indeed the case.

\subsection{Non-linear interactions}
Fig. \ref{fig:SxSSlice} is an exemplar of the inadequacy of the linear summation approach in the \emph{Leader Follower} case, showing D-axis predictions at the mean formation position for three concurrent neighbours.
In the topmost graph we see three distinct peaks, which are a result of each vehicle flying at a \SI{0.5}{m} offset from the formation centroid.
These measurements match closely with the predictions from a na\"ive linear sum model, which is able to account for most of the D-axis measurements (the largest error is less than \SI{20}{\percent} of the measured forces).
However, this begins to fail at larger vertical separations between the load-stand and formation.
%Whilst this result shows that a na\"ive numeric linear summing model can account for most d-axis force, with the largest error being less than 20\% of observed forces, the following graphs disprove this.

The lower two graphs of Fig. \ref{fig:SxSSlice} are identical in methodology to the top, excepting the altitude of the formation, which are 0.5 and 1m higher respectively. It can be seen that the measured forces concentrate away from the formation's edges with increasing altitude. Once an altitude difference of 1.3m between the sufferer and formation is reached, as in the final graph, the magnitude of the force predictions of a summing model approach the scale of those observed in total. This effect can more clearly be seen when observing a contour map of forces over the N-E plane, as in Fig. \ref{fig:NEContour}.

This example clearly demonstrates our strongest observation of a non-linearity, which we summarise as follows:

\begin{figure}[tb]
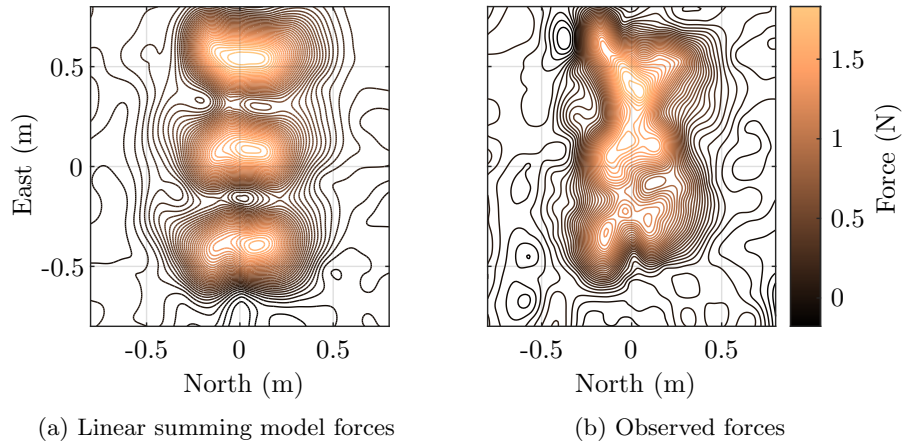

    \centering
    \begin{subfigure}[b]{0.46\textwidth}
        \centering
        \includesvg[width=\textwidth]{images/Seq3HighContourMod.svg}
        \caption{Linear summing model forces}
        \label{fig:NEContourMod}
    \end{subfigure}
    \begin{subfigure}[b]{0.52\textwidth}
        \centering
        \includesvg[width=\textwidth]{images/Seq3HighContourObs.svg}
        \caption{Observed forces}
        \label{fig:NEContourObs}
    \end{subfigure}
    \caption{ Modelled and Observed forces experienced by the load-stand from the 3-multirotor \textit{Leader Follower} case at a relative altitude of 1.3m to the formation. This is the same case as Fig. \ref{fig:SxSSlice}. We observe that, compared to a linear summing model, the observed forces are significantly less distinct for each multirotor. }
    \label{fig:NEContour}
\end{figure}

%\vspace{1em}
%\noindent\fcolorbox{black}{lightgray}{%
%\parbox{\textwidth}{%
\textbf{\centering{ The downwash airflow of multirotors in near proximity tend to merge laterally, and reduces in overall cross-sectional area with increasing distance from the generating multirotors.}}
%}%
%}
%\vspace{1em}

We make this observation consistently across multiple axes, with the impact upon Pitch and Roll torques being most evident after D-axis force. It is interesting to note in a comparison between Fig. \ref{fig:NEContourObs} and Fig. \ref{fig:NEContourMod}, the downwash of the last multirotor in the formation will appear to move forwards relative to the multirotor that produced it, demonstrating the strength of this effect.

\subsection{Non-linear models}
\begin{table}[t]
    \centering
    \caption{Prediction error for 3 formations over all 6 measurement axes. Both the Learnt Linear model predictions and Learnt Non-Linear models are included for each axis and formation. Values are the integral of the model predicted error over the experimental arena area in the lateral plane, minus the ground truth, sampled at a relative altitude of 1.3m. Lower is better, with a value of 0 matching the ground truth.}
    \begin{tabular}{|>{\centering}m{2.5cm}|>{\centering}m{2cm}|>{\centering}m{1cm}|>{\centering}m{1cm}|>{\centering}m{1cm}|>{\centering}m{1cm}|>{\centering}m{1cm}|>{\centering\arraybackslash}m{1cm}|}
        \hline
        \textbf{Formation} & \textbf{Model} & \textbf{N} & \textbf{E} & \textbf{D} & \textbf{Pitch} & \textbf{Roll} & \textbf{Yaw} \\
        \hline
        \textit{Side by Side} (K=2) & Linear Non-Linear & 0.212 \textbf{0.111} & 0.095 \textbf{0.074} & 0.311 \textbf{0.161} & \textbf{0.023} 0.026 & 0.027 \textbf{0.018} & \textbf{0.007} 0.008 \\
        \hline
        \textit{Stack} (K=2) & Linear Non-Linear & 0.150 \textbf{0.095} & 0.107 \textbf{0.074} & 0.308 \textbf{0.131} & 0.028 \textbf{0.025} & 0.020 \textbf{0.018} & \textbf{0.009} 0.016 \\
        \hline
         \textit{Leader Follower} (K=3) & Linear Non-Linear & 0.173 \textbf{0.128} & 0.135 \textbf{0.121} & 0.439 \textbf{0.255} & \textbf{0.032} 0.040 & \textbf{0.033} 0.035 & \textbf{0.009} 0.013 \\
        \hline
    \end{tabular}
    \label{tab:learntable}
\end{table}

These results address \textbf{Q2}, where we see that models with more representative power outperform those limited to linear combinations of single vehicle data.

Table \ref{tab:learntable} shows a comparison between our linear and non-linear learnt models. In general it can be seen that the non-linear model is superior at reducing the overall error between model predictions and the ground truth for many cases, particularly D-axis forces. E, Pitch, Roll and Yaw forces and torques are generally similar for both model types; this is due to the fact that, with the exception of Yaw torques, each of these axes are either dominated by noise or linearly represent-able. As neither model is superior to the other for those circumstances, their scores are similar in this benchmark. Yaw is largely dominated by noise, both in the form of chaotic interactions and due to the very low forces that appear to be exerted.

We attribute the difference between N-axis and E-axis force prediction performance to the experimental setup, specifically the fact that all trajectories are flown parallel to the E-axis, and simultaneously that our lab is substantially narrower in the N-axis. We believe these factors combine to generate non-linearities that are more difficult to represent in a learnt model.

\begin{figure}
    \centering
    \includesvg[width=1.0\linewidth]{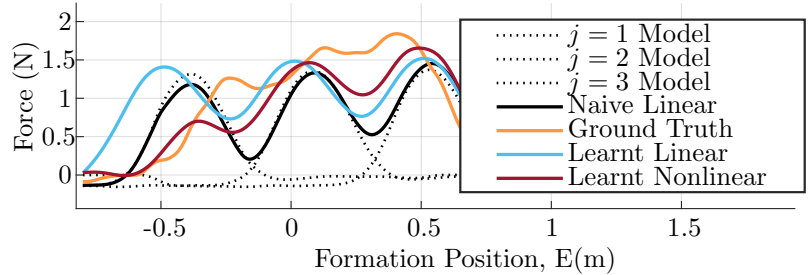}
    \caption{A comparison of all the models deployed during this work, evaluated using the $K=3$ \textit{Leader Follower} formation at 1.3m relative altitude. We observe that non-linear models can capture the proportional upwards trend that correlates the ground truth force and formation position in the E-axis.}
    \label{fig:nonlineargraph}
    \vspace{-4ex}
\end{figure}

Finally, Fig. \ref{fig:nonlineargraph} shows a direct comparison of all three evaluated models in this work for the $K=3$ \textit{Leader Follower} formation, at 1.3m relative altitude.
This figure shows the differences between our three models and the ground truth data.
The \textit{Na\"ive Linear} can be seen as a simple summation of the individual $j=1,2,3$ models, demonstrating no reaction to complex patterns. The \textit{Learnt Linear} model exhibits a similar behaviour, despite the fact that it was trained on multi-vehicle data, because its internal structure is only capable of building a single vehicle model and then querying that model multiple times for summation.
Finally, the \textit{Learnt Nonlinear} model is the only model that responds to the reduced forces for negative E-axis positions, and exhibits the lowest overall error.

%% file: 04-insights.tex
\vspace{-2ex}
\section{Discussion}
\vspace{-2ex}
Throughout this paper, we have observed several factors that are key to the understanding of multirotor downwash interactions in dense swarms:
\vspace{-1ex}
\begin{itemize}
    \item Multirotor downwash fields expand very little with vertical distance, and remain concentrated approximately within the body radius up to at least 1.5m, with the originating multirotor being $<30$cm in total diameter.
    \item The aerodynamic influence of a multirotor is negligible to its neighbours outside of its downwash (Fig. \ref{fig:SingleContour}), including at very small ($<1$ vehicle diameter) separations. Notwithstanding this, in our enclosed test space we observed lateral forces being exerted at significant distance from the platform, though these were generally chaotic and of a small magnitude. This primarily impacted our formation flights because lateral position keeping accuracy decreased with increasing $K$.
    \item Prediction errors are highly dependent upon the formations of the neighbours, with the \emph{Leader Follower} formation generally resulting in the largest errors. We posit that this scenario is more prone to stray turbulence from the lead vehicle, thus inducing non-linearity. We reinforce this hypothesis by observing a similar behaviour in the \textit{Stack} formation, which naturally puts vehicles within each other's downwash. Despite this, a smaller initial lateral distance in the stack formation reduces the observed nonlinearity versus \emph{Leader Follower} formation, because the airflow undergoes less lateral contraction.
\end{itemize}

Given the above, we reason that simple summation of forces from single vehicle models is insufficient when a sufferer is subjected to multiple downwash zones because it can result in force estimation errors up to $\approx\SI{3}{N}$ in our datasets. Given our \SI{700}{g} platform, this would result in an acceleration of $\approx\SI{4}{m/s^2}$, and represents $\approx\SI{75}{\percent}$ of peak forces out of all experiments. When combined with the sharp boundary of the downwash field, this rate of change is hard for multirotors to counteract without accurate predictive models.

\vspace{-2ex}
\section{Conclusion and Future Work}
\vspace{-2ex}
In this paper we demonstrate a method for capturing and modelling downwash forces formations of up to 4 multirotors in total, flying in close formation ($<1$ body radius separation). We present the scenarios we observed to be most likely to produce large nonlinearities, and show that deep-set based neural network models are capable of outperforming simple linear summation for force prediction.

Throughout this work, we have only analysed a limited sample of velocities. We observed during the project that lower velocities tend to exaggerate downwash effects, however a more detailed study is indicated.

Further, our deep-sets based non-linear model may be incapable of generalising to cases where $K>3$, either due to a lack of data or underlying patterns in the data that our learning methods cannot efficiently learn. Identifying methods of encoding the problem space into our model's learning process is an area that may permit efficient, large scale generalisation.

\subsubsection{Acknowledgment.}
This work was supported in part by ARL DCIST CRA W911NF-17-2-0181, the European Research Council (ERC) Project 949940 (gAIa), and by a gift from Arm. J. Gielis was supported by an EPSRC Doctoral Training studentship.
Their support is gratefully acknowledged. 